%% file: bare_conf.tex
\documentclass[a4paper,conference]{IEEEtran}
\usepackage{cite}
\usepackage{amsmath}
\usepackage{amssymb}
\usepackage{caption}
\usepackage{subcaption}
\usepackage{hyperref}
\usepackage{graphicx}
\usepackage{subfiles} 
\usepackage[sort&compress,numbers]{natbib}

\newcommand\blackautoref[1]{{\hypersetup{linkcolor=black}\autoref{#1}}}

\begin{document}
\begin{spacing}{0.9}

%

\title{FitGAN: Fit- and Shape-Realistic Generative Adversarial Networks for Fashion}


\author{\IEEEauthorblockN{Sonia Pecenakova \hspace{4mm}  Nour Karessli \hspace{4mm}  Reza Shirvany}
\IEEEauthorblockA{Zalando SE - Berlin, Germany}
{\tt\small \{firstname.lastname\}@zalando.de}
}


%


\maketitle

\begin{abstract}
\subfile{sections/00_abstract}
\end{abstract}


%
\IEEEpeerreviewmaketitle

\section{Introduction}
\subfile{sections/01_introduction}


\section{Related Works}
\subfile{sections/02_related_works}

\section{Method}\label{sec:method}
\subfile{sections/03_method}

\section{Experiments}
\subfile{sections/04_experiments}

\section{Conclusion}
\subfile{sections/05_conclusion}

\hfill \break \textit{\textbf{Acknowledgements}. The authors would like to thank \mbox{Nikolay~Jetchev} for the positive energy and scientific discussions contributing to the success of this work.}

{\small
\bibliographystyle{unsrt}
\bibliography{main,egbib}
}


\end{spacing}
\end{document}


%
\title{Supplementary Material for FitGAN: Fit- and Shape-Realistic Generative Adversarial Network for Fashion}

\author{\IEEEauthorblockN{Sonia Pecenakova \hspace{4mm}  Nour Karessli \hspace{4mm}  Reza Shirvany}\\
\IEEEauthorblockA{Zalando SE - Berlin, Germany}\\
{\tt\small \{firstname.lastname\}@zalando.de}
}


%


\maketitle

\section{Method}
We train the networks on 4 Tesla V100 GPUs using the Adam Optimizer~\cite{Kingma2014} with learning rate 0.0025 and batch size of 64. We use data augmentations such as scaling, rotations, brightness and contrast changes, following existing configurations of~\cite{karras2020training}. We use 90\% of our data for training and keep 10\% as a hold-out test set which is used for latent space exploration and evaluation experiments. The training images are downsized to 256x256 pixels, with the width padded with white pixels to achieve a squared shape. The generator architecture consists of 2 mapping layers, which maps 512-dimensional latent space $Z$ to a 512-dimensional intermediate latent space $W$, and 7 synthesis blocks, that produce image resolutions from 4x4 up to 256x256. The discriminator mirrors the architecture of the generator, with 7 convolutional blocks, including skip connections. For semi-supervised approach, both supervised and unsupervised networks share the layers with the discriminator except for the last one, which is a linear layer with the given number of outputs, in case of supervised classification followed by a softmax transformation. The supervised conditioning labels are one-hot-encoded. We use 1 for the supervised loss term weight $\gamma$ and 0.1 for the unsupervised loss term weight $\beta$.

\section{Experiments}
\subsection{Dataset}
\begin{figure*}[b!]
\begin{center}
  \includegraphics[width=\linewidth]{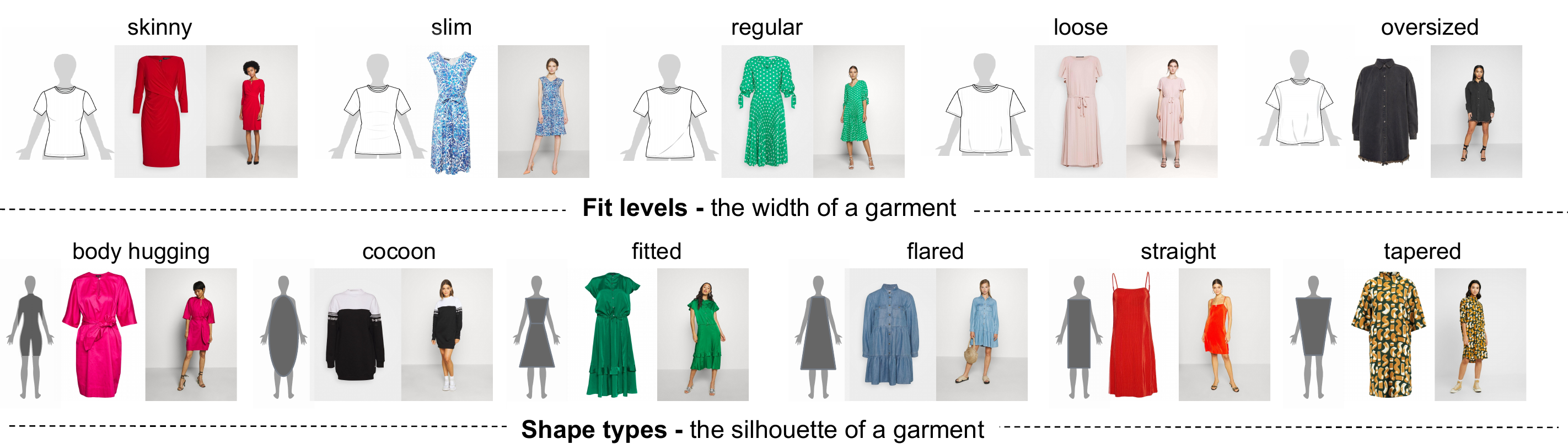}
\end{center}
    \caption{\textbf{Fit and shape of a garment}. The fit describes intended distance from the body whereas shape describes the silhouette. For each level/type we show a reference illustration used for annotation guidance (left), an example garment packshot image (middle), and model image (right). }
\label{fig:uft_data}
\end{figure*}

Our dataset consists of packshot and model images described with their fit and shape attributes. The fit of an article describes the article's intended distance from the body and can be ordered into 5 levels, from closest to body to furthest. The shape of an article describe the silhouette or the cut and can be one of 6 defined categories. You can see an example of each of these attributes in~\blackautoref{fig:uft_data}. In \blackautoref{tab:data_size} we also provide detailed number of samples per class combination. The samples are unique articles, which can have packshot and/or model image.

\begin{table}[t!]
\caption{\label{tab:data_size}\textbf{Number of samples in dataset.} One of the main challenges of the dataset is the large class imbalance.}
\begin{center}
\begin{tabular}{l|rrrrr}
\multicolumn{6}{c}{\textbf{Women's Dresses}} \\
\hline
\textbf{Shape $\downarrow$  Fit $\rightarrow$}  &  Skinny &   Slim &  Regular &   Loose &     Oversized \\
\hline
Body-Hugging &          1,961 &     278 &   16 &        2 &         0 \\
Cocoon &                2 &         16 &    82 &	    266 &   	113\\
Fitted &                2,066 &	    6,040 & 8,612 &     920 & 	    17 \\
Flared &                26 &        551 &   2,979 &     3,086 &     122 \\ 
Straight &              10 &        269 &   4,638 &     2,019 &     213 \\
Tapered &               11 &        48 &    57 &        32 &        9 \\
\hline
\multicolumn{6}{c}{} \\
\multicolumn{6}{c}{\textbf{Women's Jeans}} \\
\hline
\textbf{Shape $\downarrow$  Fit $\rightarrow$}  &  Skinny &   Slim &  Regular &   Loose &     Oversized \\
\hline
Body-Hugging &          6,419 &     230 &   8 &         1 &         0 \\
Cocoon &                3 &         3 &     5 &	        128 &   	0\\
Fitted &                923 &	    1,187 & 32 &        64 & 	    0 \\
Flared &                149 &       475 &   263 &       485 &       0 \\ 
Straight &              90 &        804 &   1,059 &     1,006 &     0 \\
Tapered &               232 &       570 &   188 &       1,226 &     0 \\
\hline
\end{tabular}
\end{center}
\end{table}

\subsection{Conditional Generation}

\begin{figure}[htbp]
\begin{center}
  \includegraphics[width=1.0\linewidth]{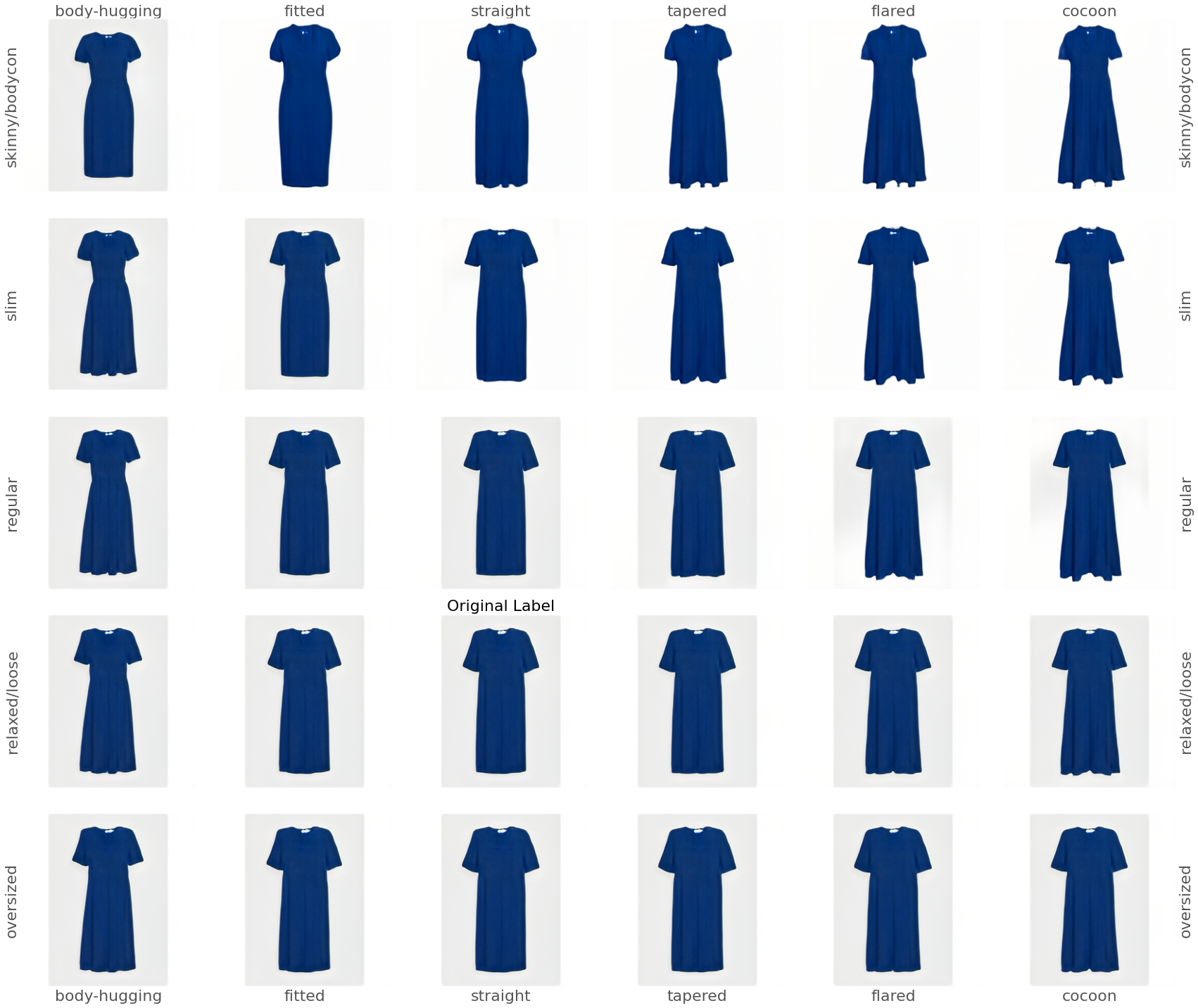}
\end{center}
    \caption{\textbf{Conditional fit and shape generation.} Images generated by taking a projection of a real image and conditioning it on all possible fit and shape combinations for Women's Dresses.}
    \label{fig:conditional_dress_fit_shape}
\end{figure}

We provide more examples of conditional image generation, taking as input projection of existing image into the latent space, and conditioning it on possible fit and shape combinations in \blackautoref{fig:conditional_dress_fit_shape}.

\begin{figure}[bth]
\centering
  \includegraphics[width=1.0\linewidth]{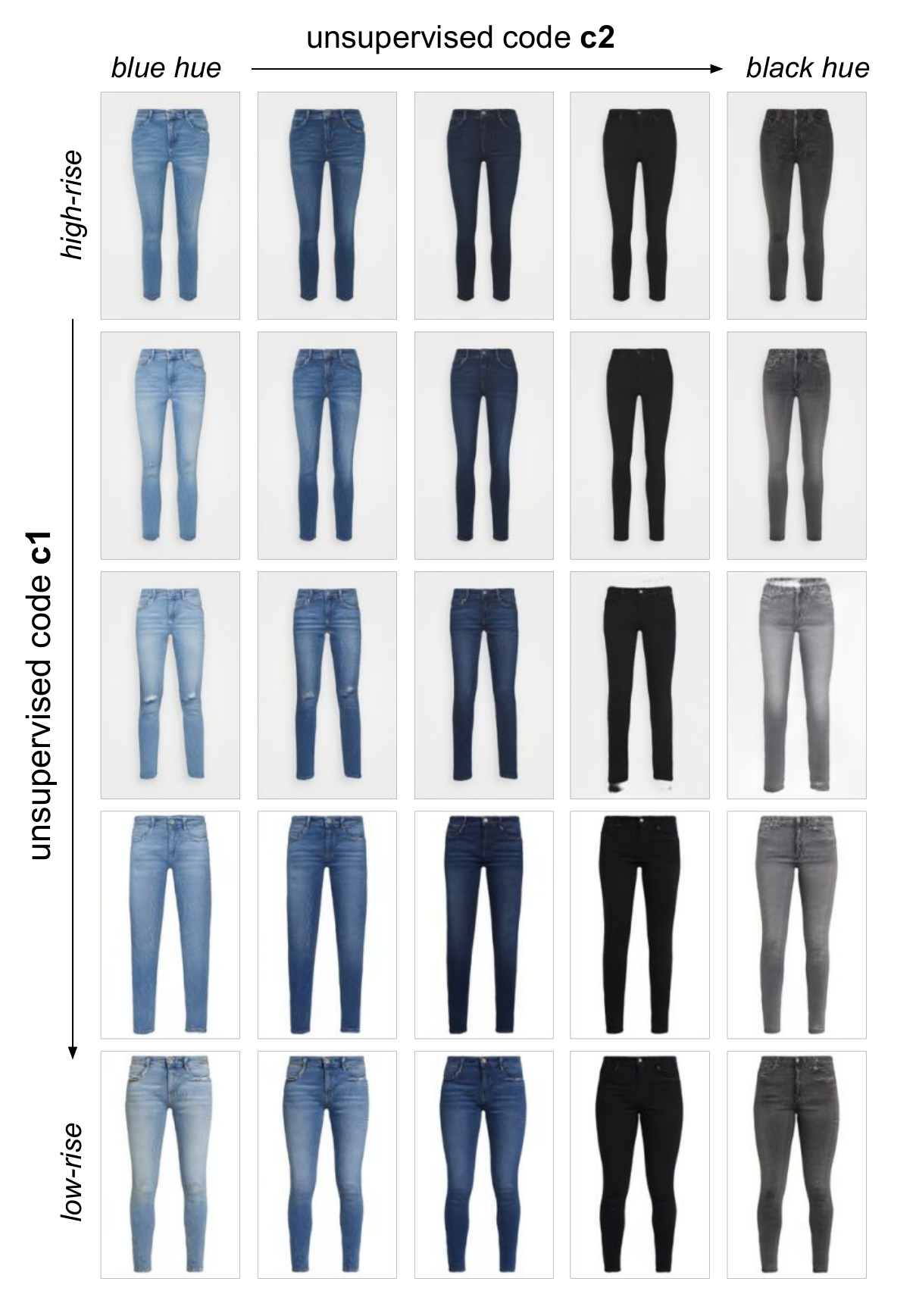}
  \caption{\textbf{Semi-supervised approach.} The generated jean in this example is conditioned to be of shape body-hugging through a categorical input to the generator. The x-axis shows a variation in the continuous code $c2$ which seems to make the jeans more blue or more black. On the other hand, variation in code $c1$ seems to represent change from high-rise to low-rise jeans. We note, that the body-hugging shape of the jeans stays constant throughout these variations.}
\label{fig:semi-supervised-jeans}
\end{figure}

\pagebreak
\subsection{Latent Space Analysis}
We study how the fit information is encoded in the learnt intermediate latent space \textbf{W} of the GAN. We compute an average class representation for each fit and shape and compare the euclidean distances between these vectors. ~\blackautoref{tab:mean_distances} shows the distance of each class $c$ to all others for both fit and shape attributes. We can see that the fit levels are encoded semantically meaningfully with Skinny fit being the closest to Slim and furthest away from Oversized fit. On the other hand, in a network that was conditioned only on one-hot-encoded shape attribute, each class representation is encoded at approximately the same distance from all others.

\begin{table}[b!]
\caption{\label{tab:mean_distances}\textbf{Euclidean distance of mean class latent vectors.} We measure the distance from each class average latent vector to other classes. We can see that the fit levels are encoded semantically meaningfully, with tighter fits like Skinny and Slim closer to each other than to Loose and Oversized fit. For shape, which does not have an implicit order, the distances are similar for all classes. Some shape classes with are omitted, as they have no or little representation in the dataset consisting of Women's Jeans.}
\begin{center}
\begin{tabular}{l|rrrrr}
\multicolumn{6}{c}{\textbf{Fit Levels}} \\
\hline
Distance &   Skinny &   Slim &    Regular &   Loose & Oversized \\
\hline
Skinny &   0 &         2.7 &       3.9 &       4.4 &  5.3\\
Slim &    - &         0 &         2.8 &      3.6 &	4.5 \\
Regular &   - &         - &	    0 &         2.2 &  3.4 \\
Loose &     - &         - &      - &       0 &     3.2\\ 
Oversized & - &       - &       - &      - &   0\\
\hline
\multicolumn{6}{c}{} \\
\multicolumn{6}{c}{\textbf{Shape Attributes}} \\
\hline
Distance &   Body-Hugging &   Straight &    Tapered &   Flared \\
\hline
Body-Hugging &   0 &         5.8 &       5.7 &       5.7\\
Straight &    - &         0 &         4.7 &      4.7 \\
Tapered &   - &         - &	    0 &         4.9 \\
Flared &     - &         - &      - &       0\\ 
\hline
\end{tabular}
\end{center}
\end{table}

\begin{figure}[htbp]
\begin{center}
  \includegraphics[width=1.0\linewidth]{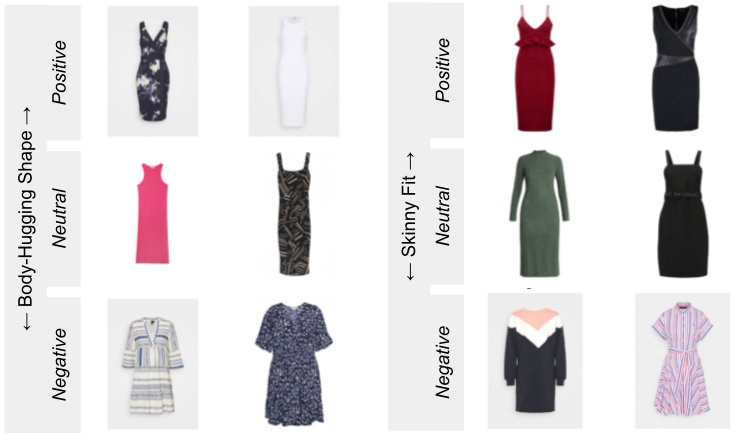}
\end{center}
  \caption{\textbf{Real images projections categorized by distances to the decision boundary of a linear classifier}. We select a few examples of images that are the furthest to each side of the decision boundary of a linear classifier for skinny fit and body-hugging shape.}
\label{fig:latent_space_svm}
\end{figure}

We further investigate how the latent space is organized by fitting a linear classifier in the latent space \textbf{W} to learn a decision boundary for each fit and for each shape. The linear SVM is trained on the latent projections of a small subset of real images that the generative model has not seen during training and evaluated on a small hold-out test. \blackautoref{fig:latent_space_svm} shows a few test examples categorized based on their distance from the decision boundary for body-hugging shape and skinny fit respectively. As we can see in the example of skinny fit, skinny articles have the largest positive distance to the boundary, whereas articles that are closer to the boundary tend to have more regular fit. Once we move away from the boundary in negative direction, more loose articles appear. This suggests that we can find linear boundaries in the latent space for attributes of interest, and use such boundaries in the future to find articles in a desired fit and shape and recommend them to user, based on purely visual features of the article's image.

\subsection{Real Image Projection}
We have previosly shown, that projecting a real image into the latent space and changing its fit and shape leads to visible and realistic fit and shape changes. However, real images projected to the latent space and subsequently reconstructed by the generator back to the image space lose some of their details, such as intricate sleeves structure and belt. This suggests that the latent space lacks the ability to encode such details and the generator collapses to a more common node with simple sleeves. 

\begin{figure*}[t!]
\begin{center}
  \includegraphics[width=0.9\linewidth]{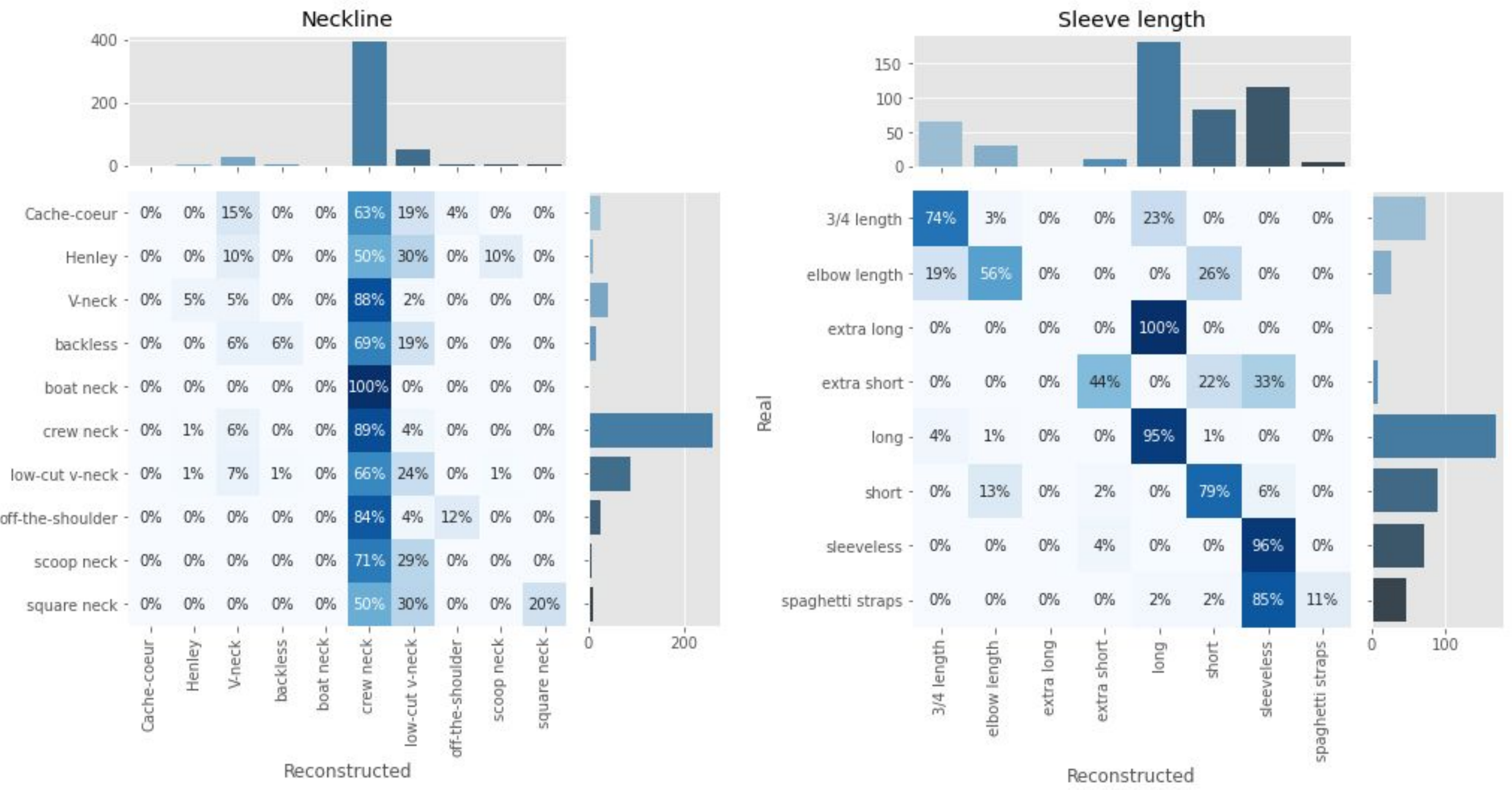}
\end{center}
   \caption{\textbf{Classification of neckline and sleeve length for real and reconstructed images}. Images from the test set are projected into the latent space $W$ and reconstructed via the generator. We see a normalized confusion matrix of a pre-trained classifier prediction for neckline and sleeve length between real and reconstructed images.}
\label{fig:fdna_evaluation}
\end{figure*}

To verify this hypothesis, we utilize rich article representations obtained from Fashion DNA~\cite{Bracher2016}, a convolutional neural network pretrained on a large-scale dataset of fashion items to predict hundreds of categorical visual attributes such as color, material, season, etc. For our purpose, we focus on the main attributes that influence the article style, namely: color, pattern, sleeve length, length, and neckline.~\blackautoref{fig:fdna_evaluation} shows a normalized confusion matrix that compare the prediction of the Fashion DNA classifier on real test images and their reconstructed versions for selected attributes. We observe that while sleeve length is reconstructed truthfully, neckline is mostly reconstructed as U-shaped neckline, which is the most common type in the dataset. The latent space struggles in encoding less common types of neckline.

\pagebreak
\subsection{Evaluation}
For domain expert annotation, we use our model D-P-F+S to generate packshot images of dresses conditioned on different fits and shapes combinations based on the fit-shape class distribution in the original training dataset
 and combined them with 15\% of real images. We opt for this split of 85-15 due to the time and resource constraints of the domain expert, as our main goal is to gather data about the generated images. The images are selected randomly and presented to the annotator in a random order. We use accuracy between the domain expert's annotations (ground truth) and the target generator condition in order to evaluate the truthful visualization of fit and shape.
 
\begin{figure}[htbp]
     \centering
     \begin{subfigure}{0.49\textwidth}
         \centering
         \includegraphics[width=\linewidth]{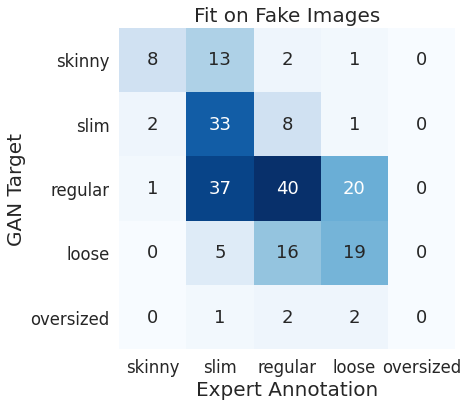}
     \end{subfigure}
     \hfill
     \begin{subfigure}{0.49\textwidth}
         \centering
         \includegraphics[width=\linewidth]{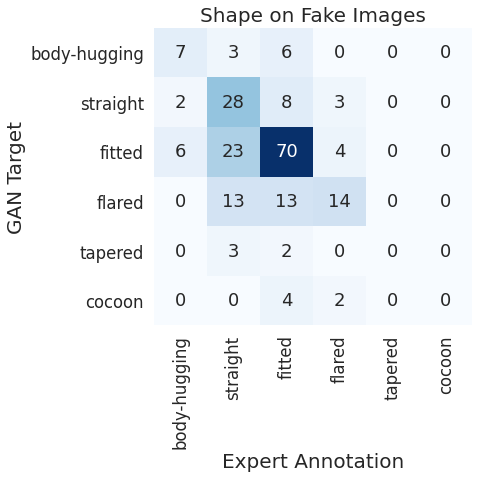}
     \end{subfigure}
        \caption{\textbf{Confusion matrix for expert annotated fit and shapes.} The confusion matrix shows the target fit and shape of images generated using FitGAN (D-P-F+S) and their corresponding annotations by the domain expert.}
    \label{fig:human_annotation_confusion_matrix}
\end{figure}

We provide detailed results in form of a confusion matrix between the target fit and shape of images generated by FitGAN and the annotations of the domain expert. \blackautoref{fig:human_annotation_confusion_matrix} shows that for fit the most differences between target and annotation are in the fit's neighboring area. We can also see that rarer shapes, such as cocoon and tapered shape are harder for the GAN to realistically generate.

\begin{figure}[t!]
\begin{center}
  \subcaptionbox{High Real Score ($\geq$ 4)}{
  \includegraphics[width=0.98\linewidth]{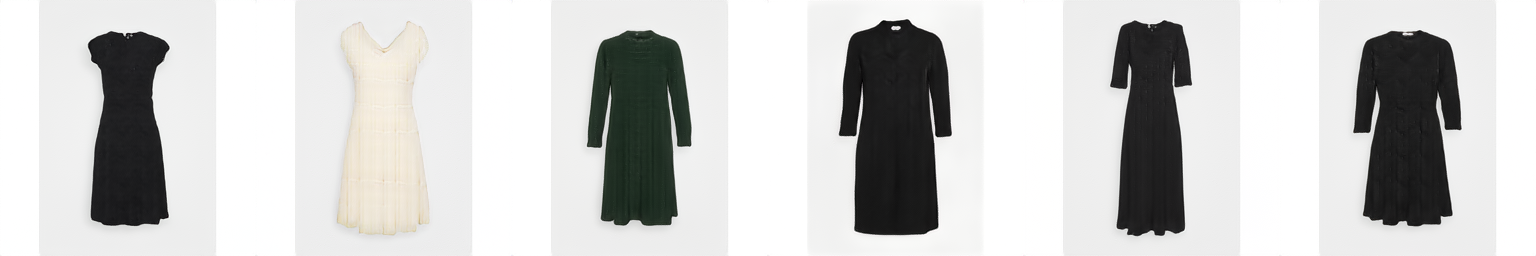}
  }
  \subcaptionbox{Low Real Score ($\leq$ 2)}{
  \includegraphics[width=0.98\linewidth]{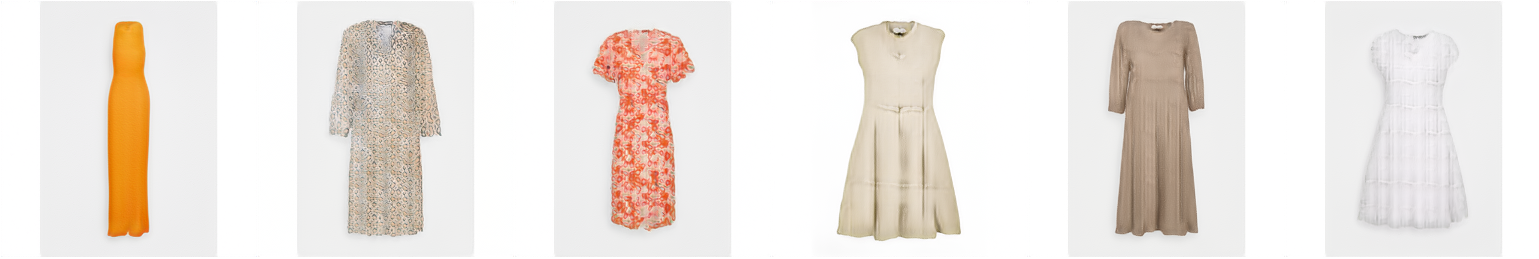}
  }
\end{center}

  \caption{\textbf{Randomly selected fake images generated by FitGAN with highest (top) and lowest (bottom) realness score}. The score is based on human annotation on a scale from 1 (not real at all) to 5 (absolutely real).}
\label{fig:human_annotation_real_fake_images}
\end{figure}

We also ask our domain expert to also rate how realistic an image is on a scale from 1 (not realistic at all) to 5 (very realistic). ~\blackautoref{fig:human_annotation_real_fake_images} shows examples of generated images with high and low realistic scores. We can see the least realistic images usually contain pattern artifacts or vague contours, whereas highly realistic images are usually black with a well-defined contour.\\

{\small
\bibliographystyle{unsrt}
\bibliography{main,egbib}
}


%% file: sections/00_abstract.tex
Amidst the rapid growth of fashion e-commerce, remote fitting of fashion articles remains a complex and challenging problem and a main driver of customers’ frustration. Despite the recent advances in 3D virtual try-on solutions, such approaches still remain limited to a very narrow -- if not only a handful -- selection of articles, and often for only one size of those fashion items. Other state-of-the-art approaches that aim to support customers find what fits them online mostly require a high level of customer engagement and privacy-sensitive data (such as height, weight, age, gender, belly shape, etc.), or alternatively need images of customers' bodies in tight clothing. They also often lack the ability to produce fit and shape aware visual guidance at scale, coming up short by simply advising which \textit{size} to order that would best match a customer's physical body attributes, without providing any information on how the garment may fit and look. Contributing towards taking a leap forward and surpassing the limitations of current approaches, we present FitGAN, a generative adversarial model that explicitly accounts for garments’ entangled size and fit characteristics of online fashion at scale. Conditioned on the fit and shape of the articles, our model learns disentangled item representations and generates realistic images reflecting the true fit and shape properties of fashion articles. Through experiments on real world data at scale, we demonstrate how our approach is capable of synthesizing visually realistic and diverse fits of fashion items and explore its ability to control fit and shape of images for thousands of online garments.

%% file: sections/01_introduction.tex
Online fashion shopping plays a prime role in many people's lives today. Virtual stores enable consumers to conveniently shop from home and quickly browse a large assortment of articles. Fashion is indeed an instant language and a strong way of self-expression. Wearing clothes allows individuals to express who they are or feel to be. Throughout our diverse cultural heritage and modern world, a person’s attire 
remains an essential code to their culture, class, personality, diversity, and even faith. In fact, the formality of clothing might not only influence the way others perceive a person, but most importantly on how people perceive themselves.

In online fashion, however, the absence of feel and touch and physical examination of items creates major hurdles as for fashion it is especially crucial for consumers to get a trustworthy feeling of how the garment would fit and look on them; even more so for articles coming in various sizes, fits, shapes, and brand specific characteristics, making it desperately hard for a customer to determine the right size and fit for their bodies and personal preferences. To alleviate this problem, online fashion platforms offer various product information including categorical features such as fabric, occasion, length, etc. and visual imagery or video content in different setups. Studies have shown that product visual information (images/videos) plays a vital role in customer's confidence, click-through rate, and purchase decision~\cite{di2014picture,goswami2011study}. However, due to the high costs in creating quality professional content, product images are often produced for only one single product size that flatters or best fits the fashion model who is wearing it. Alas, often online model body shapes and what fits them best doesn't necessary resemble customers’ diverse body shapes and fit preferences. Within this context, being able to see a garment in a different size, fit on a different body shape has game-changing value in supporting customers to understand how the garment would fit and look on them and which size and fit might be the best choice for them.

From a different perspective, recent work has focused on predicting fashion trends by collecting data from fashion weblogs, news sites and magazines, in order to identify actual and future design trends~\cite{beheshti2015trendfashion}. Being able to visually control a garment's fit and shape at scale and with low cost creates significant possibilities not only for improving the size and fit characteristics of current designs but also in developing novel fashion that is guaranteed to fit customers at scale.

To that end, in this work we propose a generative framework for realistically conveying physical fit and shape attributes of garments with minimal conditioning, a first in the fashion domain to the best of our knowledge. We denote our approach FitGAN and demonstrate through extensive experimentation with challenging real life data a fundamental baseline that is capable of achieving controllable generation and rich semantically meaningful article representations, disentangling fit and shape properties of complex garments. This work paves the way towards truthfully visualizing size and fit characteristics of fashion articles at scale with low cost (compared to physical simulations and 3D scanning) and can potentially enable fashion designers to visually control fit and shape of garments and create novel fashion designs with algorithmic assistance. As a first step we showcase that we can identify articles in similar fit and shape based on their visual image representation, which can be used for fashion recommendation based on fit and shape preferences.


The contributions of this work are 3-fold:
\vspace{-0.1cm}
\begin{itemize}
\item[1.] We introduce FitGAN, a generative framework to the fashion domain capable of realistically  incorporating the intricate physical fit and shape characteristics of garments at scale with minimal conditioning -- a first to our best knowledge; 
\item[2.] We learn a latent space that is able to disentangle the fit and shape of a garment based on its representation as 2D image. We thoroughly study the class latent representation and show that our method is able to infer the implicit semantic order of the garments' fit. We also employ rich article representations to investigate the underlying correlations between fit characteristics and other article attributes;
\item[3.] We extensively evaluate not only the quality and variety of the generated outputs using well-known GAN metrics, but also how truthfully an article's fit and shape characteristics are visualized by our method in two steps. First we leverage a supervised auxiliary model for large-scale evaluation. Then we complement it with a high-fidelity human evaluation, leveraging the knowledge of a domain expert on a small-scale test set.
\end{itemize}

%% file: sections/02_related_works.tex
Recently, an emerging body of research addressing the challenging sizing problem in online fashion has been established. Different approaches offer a personalized size recommendation (e.g. recommending size M for shirts) based on customer orders and returns data~\cite{Sembium2017,Sembium2018,Guigoures2018,Sheikh2019,Dogani2019,Lasserre2020,Hajjar2020} or personal data such as age, height, weight, and less privacy-sensitive data like a reference article~\cite{Yuan2019,Januszkiewicz2017,Lefakis2020}. Other methods provide article based size advice (e.g. recommending one size smaller than the usual) leveraging orders and returns data ~\cite{Nestler2020, Baier2019} or article images and attributes~\cite{Karessli2019}. Further enriching the image based approach of~\cite{Karessli2019}, ~\cite{hsiao2020vibe} introduced a body-aware visual embedding that captures clothing’s affinity focusing on different body shapes. 

\begin{figure}[b!]
\centering
  \includegraphics[width=0.95\linewidth]{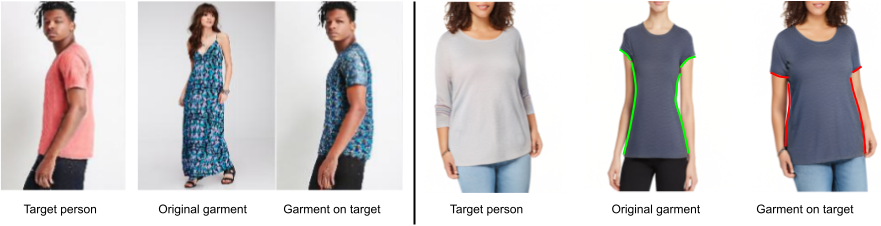}
  \caption{\textbf{Limitations of existing approaches}. Left: ADGAN~\cite{men2020controllable} fusing original garment with target person outfit into a physically non-existing garment. Right: TryOnGAN~\cite{lewis2021tryongan} altering original garment's physically designed fit from slim to an unrealistic loose fit for visual appealing.}
\label{fig:limitations}
\end{figure}

Other line of research leverages 3D models and high fidelity garment data to render garments on various body shapes with high level of accuracy~\cite{bhatnagar2019multi, Thalmann2011, Surville2013, Peng2014, Januszkiewicz2017, patel2020tailornet}. However, such approaches mostly depend on accurate 3D scans or digitized production design data that are costly or challenging to acquire. 

In the broader computer vision community, Generative Adversarial Networks (GANs)\cite{goodfellow2014generative} have become the state-of-the-art technique for producing photo-realistic images. 
Recent advances, most notably StyleGAN~\cite{karras2019style,karras2020analyzing} have achieved impressive high resolution results by redesigning the generator architecture. Other field of generative models, that has recently seen a lot of advances and promises more stable training, are diffusion models~\cite{song2019generative,ho2020denoising,nichol2021improved}. 

Many recent works using generative models in the context of fashion focus on what is loosely called virtual try-on; given an image of a garment and a target person, the garment is visualized wrapped on the target person with high visual appeal and realism, not necessarily guaranteeing the physical feasibility in the real world. A virtual try-on method using conditional GANs~\cite{mirza2014conditional} eliminating the dependency on 3D datasets, was first introduced in Viton\cite{han2018viton}. 
\mbox{Outfit-VITON}~\cite{neuberger2020image} further overcomes the need for paired dataset and produces higher resolution and quality images, consisting of an entire outfit of multiple reference garments. \mbox{Tryongan}~\cite{lewis2021tryongan} continued the work with unpaired data, while preserving the identity and body shape of the target person, while ADGAN~\cite{men2020controllable} replaces the need of paired dataset with images of persons wearing the same garment in multiple poses, composing the generated result from various input sources. Most of the aforementioned approaches add a segmentation branch to enable wrapping the garment more realistically. The \mbox{Outfit Renderer}~\cite{yildirim2019generating} instead conditions the GAN on an embedding of multiple garments and a target pose producing realistic images of models wearing entire outfits without controlling the model's identity or body shape.  

Despite achieving highly appealing visual representations, none of these methods are currently able to account for the fit characteristics of the garment, and thus, one can easily end up with visualizations that are impossible in real world where none of the (limited) available size would physically fit as visualized. Although~\cite{lewis2021tryongan,men2020controllable} try to implicitly preserve the body shape of the model, by using segmentation and pose keypoints, whether the wrapping of the garment and its fit behaviour is real-world realistic is not evaluated. ~\blackautoref{fig:limitations} illustrates examples of limitations of ADGAN~\cite{men2020controllable} and TryOnGAN~\cite{lewis2021tryongan} in fit and shape domain.

%% file: sections/03_method.tex
\begin{figure}[t!]
\centering
  \includegraphics[width=1.0\linewidth]{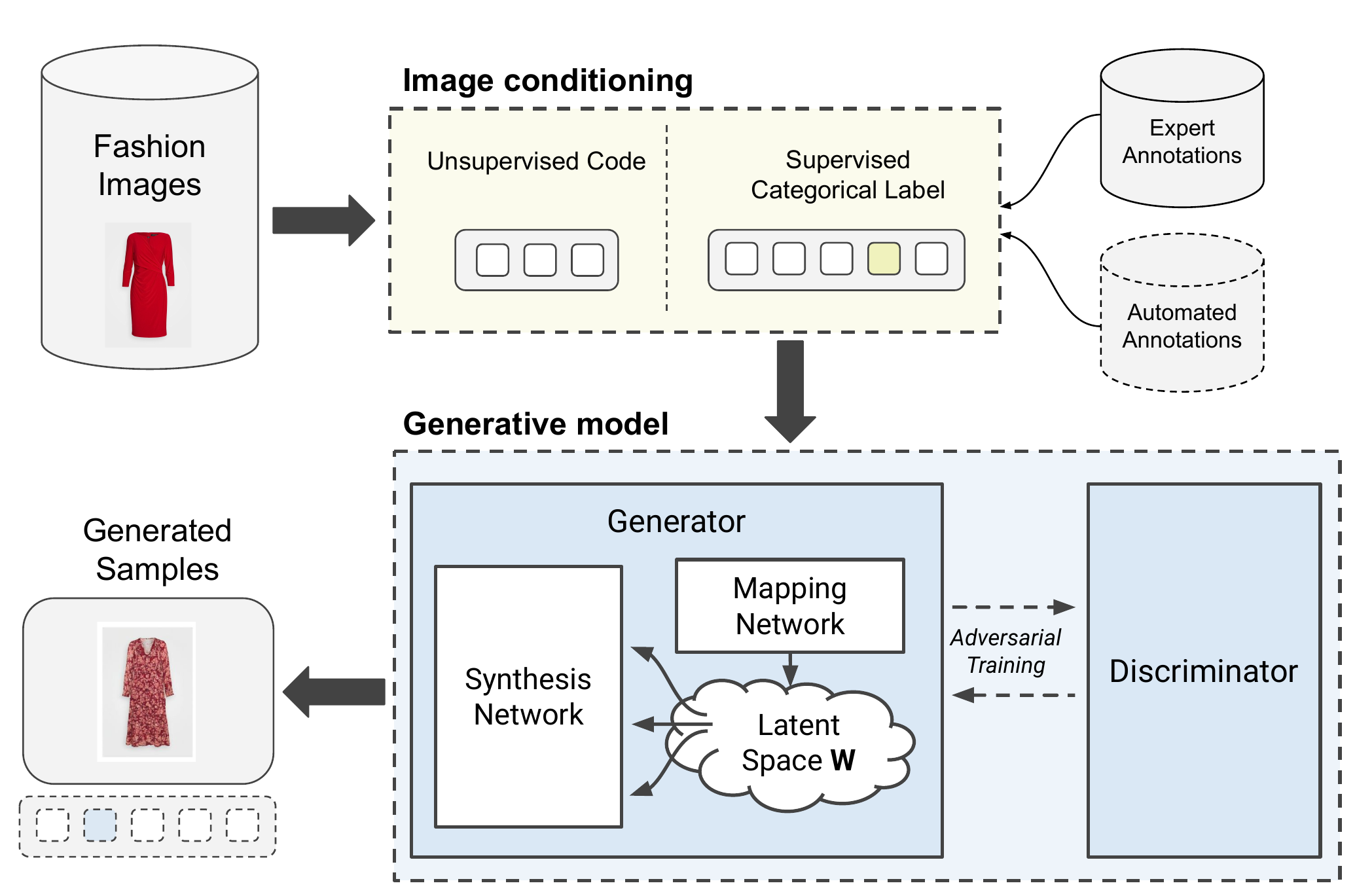}
   \caption{\textbf{FitGAN framework}. Given a dataset of fashion images, we enrich them with supervised and/or unsupervised conditioning and train a generative model to generate new samples from the dataset controllable by the conditioning label.}
   \vspace{-0.2cm}
\label{fig:architecture}
\end{figure}

In this section we describe the proposed generative framework
for realistically conveying physical fit and shape attributes of garments with flexible backbone and conditionings. Given a dataset of real images, our objective is therefore to be able to sample new examples from this data distribution for which we can control the fit and shape characteristics of the sampled garments.~\autoref{fig:architecture} provides a high level overview of the architecture of the method. In what follows, we first discuss the possible backbones for the generative module of the conditional framework; next we discuss how to gain control of the fit attributes through supervised and semi-supervised conditioning approaches; and lastly, we introduce an evaluation method to assess the quality of the generated samples in respect to their fit characteristics.

\subsection{Generative model backbone}
\begin{figure}[b!]
\begin{center}
  \includegraphics[width=0.85\linewidth]{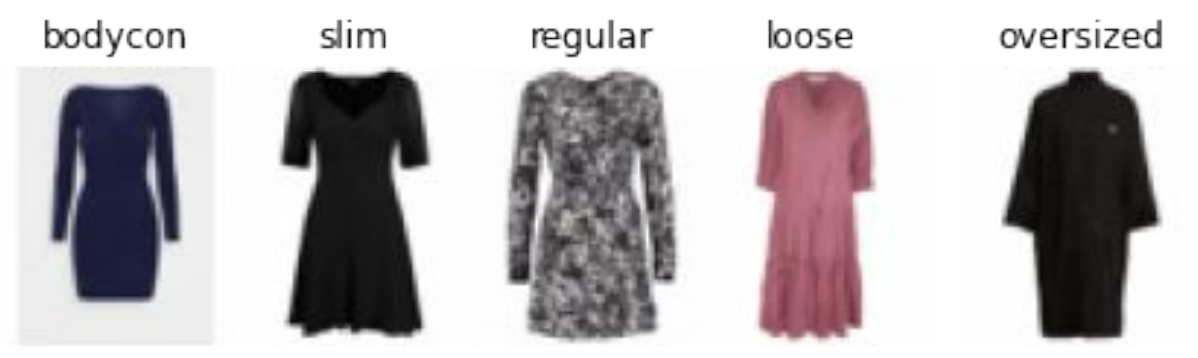}
\end{center}
    \caption{\textbf{Generated samples using a diffusion backbone conditioned on fit.}}
\label{fig:diffusion}
\end{figure}

The backbone of the proposed framework, a generative model that learns the underlying image distribution and can sample new images conditioned on fit attributes, is by design independent from any specific or custom-made model and can be flexibly swapped using any state-of-the-art models. For the work in hand, we have explored two types of generative models as backbone- generative adversarial networks (GANs)~\cite{goodfellow2014generative} and diffusion models~\cite{song2019generative}. A generative adversarial network consists of two networks playing the minimax game, whereas a diffusion model gradually adds random noise to data and subsequently learns to reverse the process. Diffusion models have the advantage of a more stable training process as they do not depend on the fickle balance between two adversarial opponents. However, the training process can be both time and resource-consuming, especially for high-resolution images. We demonstrate some lower-resolution samples generated by a diffusion model~\cite{nichol2021improved} in~\blackautoref{fig:diffusion}. Here, we focus on the established GANs, while monitoring future advances in the field of diffusion models.

In a generative adversarial network, the generator $G$ tries to learn the underlying data distribution $p(x)$, mapping a random input vector $z$ to the image space $\hat{p}(x)$. The discriminator $D$, given an image $x$, predicts whether it comes from the real distribution $p(x)$ or fake distribution $\hat{p}(x)$. Discriminator is trained to maximize the probability of classifying the sample distribution correctly, while generator wants to minimize this probability. The generator and discriminator parameters are optimized to maximize the non-saturating logistic loss:

\begin{center}
$
\mathcal{L}_{GAN} =  E_{x}[\log D(x)] + E_{z}[\log (1 - D(G(z)))]
$
\end{center}

Most recently, the StyleGAN architecture~\cite{karras2019style} has shown impressive high-resolution results on datasets such as human faces\cite{karras2017progressive}. Given the challenges of our problem, discussed in detail
in~\blackautoref{sec:dataset}, and the potential of StyleGAN-ADA~\cite{karras2020training}we use the latter as one of the promising backbones. StyleGAN-ADA~\cite{karras2020training}uses additional adaptive distriminator augmentation to stabilize the training with limited data and is based on StyleGAN2 architecture\cite{karras2020analyzing}. As shown in~\autoref{fig:architecture}, the generator consists of an additional mapping network, which maps the random latent vector $z$ to an intermediate latent space \textbf{W}. The sampled intermediate latent vector $w$, also called \textit{style code} is then injected into the synthesis network at multiple points, using Adaptive Instance Normalization (AdaIN)~\cite{huang2017arbitrary}. This architecture not only results in high image quality, but also provides latent space disentanglement qualities and style-mixing abilities that further benefit our approach, which we explore in~\blackautoref{sec:latent_space}.

\subsection{Conditioned generation}
To control the output of the generative model with respect to fit and shape of the generated garment, we integrate an additional conditioning of the network to the training process. We develop a conditional generative network\cite{mirza2014conditional}, where an image $x$ is paired with a conditioning label $c$. The conditioning label is provided as input to the generator together with the random input $z$ so that $G(z, c) \xrightarrow{} \hat{p}(x)$. To force the generator to consider the conditioning label, the conditioning label is also provided to the discriminator as an additional input paired with real and fake images so that $D(x,c) \rightarrow y$. The composition of the conditioning label is flexible based on the specific task at hand.
\\
\textbf{Supervised Conditioning:} In order to establish a comprehensive baseline in this domain, we first consider a supervised conditioning approach, where the condition $c$ comes from an annotated dataset, consisting of pairs of real images and their fit and shape attributes. As shown in~\autoref{fig:architecture}, this approach can be extended by using labels from other sources as well. In this case, the label is represented as a one-hot-encoded vector and fed into each the generator and discriminator via linear layers. The discriminator is trained to maximize, while generator wants to minimize the loss:

\begin{center}
$
\mathcal{L}_{CGAN} =  E_{x}[\log D(x, c)] + E_{z}[\log (1 - D(G(z, c)))].
$
\end{center}

\textbf{Semi Supervised Conditioning:} The categorical fit and shape labels describe the fit and shape characteristics of a garment, but there are many more relevant aspects to the garments, such as color, pattern, length, sleeve length etc. the labelling of which would be laborious and amplify class imbalance. In order to capture these attributes of garments without the need for explicit supervision, we explore a semi-supervised conditioning approach, inspired by the advances in Info-StyleGAN~\cite{nie2020semi}. The previously introduced supervised categorical label $c_{s}$, is concatenated with an unsupervised code $c_{u}$ sampled from a uniform distribution $\mathcal {U}_{[-1,1]}$. The dimensionality of the unsupervised code and its sampling distribution can be chosen freely.

The discriminator $D$ no longer receives the conditioning label as input but is instead trained to predict it. Additionally to the network $D$, which also predicts whether an image is real or fake, we train a supervised classifier $C$  with a softmax output to predict the supervised categorical label $c_{s}$, and an encoder network $E$ with a linear output to predict the continuous conditioning code $c_u$. Both $C$ and $E$ networks share all layers with the discriminator $D$ except for the last one. 

The supervised classifier $C$ has two objectives $\mathcal{L}_C^r$ and $\mathcal{L}_C^f$. Given a pair of real image and its ground-truth label $(x, c_s)$, it is trained to classify the image correctly, whereas the generator is optimized to fool the classifier so that a fake image $\hat{x}$ is classified according to its target conditioning label $c_{\hat{s}}$:

\begin{center}
$
\mathcal{L}_C^r = \mathbb{E}_{x, c_s}[-\log C(c_s|x)]
$
$
\mathcal{L}_C^f = \mathbb{E}_{z, c_{\hat{s}}}[-\log C(c_{\hat{s}}|G(z, c_{\hat{s}})].
$
\end{center}

The encoder $E$ is trained to correctly reconstruct the unsupervised code $c_u$, given a fake image $\hat{x}=G(z,c_u)$:

\begin{center}
$
\mathcal{L}_E = \mathbb{E}_{z \sim p(z),c_u \sim \mathcal {U}_{[-1,1]}} [||c_u - E(G(z,c_u)||_{2}].
$
\end{center}

The final objective of the networks if as follows:

\begin{center}
$
\mathcal{L}^{(G)} = \mathcal{L}_{CGAN} + \gamma \mathcal{L}_C^f + \beta \mathcal{L}_E
$

$
\mathcal{L}^{(D)} = - \mathcal{L}_{CGAN} +  \gamma \mathcal{L}_C^r,
$

\end{center}

where $\gamma$ and $\beta$ are hyperparameters, which control the trade-off between supervision and the generative objective. For exact training details and hyperparameters, please refer to the Supplementary Materials.

\subsection{Evaluation model}
\label{sec:evaluation-model}
As the main objective of our method is to realistically capture fit and shape characteristics of garments, we train an auxiliary network to evaluate how well the generated images are able to represent fit and shape attributes. The evaluation network is a supervised classification network, which follows the ResNet50~\cite{he2016deep} architecture, and is trained on the same dataset as the generative network to classify fit and shape of a garment, given an image of the article and its ground truth label. In order to achieve satisfactory performance with limited training data, we initialize the network with pre-trained weights from a fashion article classification task~\cite{Bracher2016}. The pre-trained layers are followed by a  multilayer perceptron with one softmax output layer to classify fit and one to classify shape of the garment. The multilayer perceptron is optimized using Adam optimizer to minimize the cross entropy loss between predicted fit and shape and the corresponding ground truth.

%% file: sections/04_experiments.tex
\subsection{Dataset}
\label{sec:dataset}

\begin{figure}[t!]
\begin{center}
  \includegraphics[width=0.85\linewidth]{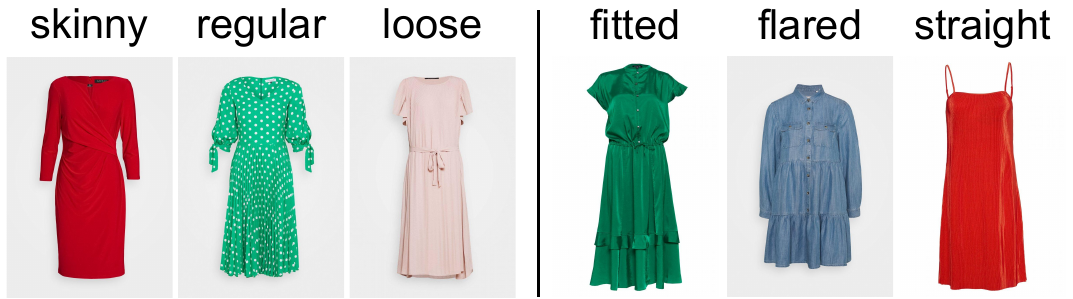}
\end{center}
    \caption{\textbf{Fit and shape of a garment}. We show a few selected fit (left) and shape (right) categories.}
    \vspace{-0.4cm}
\label{fig:uft_data}
\end{figure}

We conducted our experiments with a dataset consisting of article images and their corresponding fit and shape descriptions. We focus on the categories of female dresses and jeans and collected around 35.000 and 15.000 articles respectively for each. Each article is pictured as packshot and/or on human model.
For detailed number of samples in the dataset, please refer to the Supplementary Materials. The dataset represent real-world cases from a major online e-commerce platform. The articles are annotated by a pool of regularly trained human semi-experts with domain knowledge and the quality of the data is assured with multiple automated checks.

Inspired by~\cite{fashionpedia,feyerabend2009fashion}, experts in fashion design have defined high level types of fit and shape that generalize to different categories of garments. The fit of an article describes the article's intended distance from the body and can be ordered into 5 levels, from closest to body to furthest. The shape of an article describe the silhouette or the cut and can be one of 6 defined categories. You can see an example of some of these attributes in~\blackautoref{fig:uft_data}. Please refer to Supplementary Materials for more examples.

As the dataset has been collected from an online platform, where articles are pictured on professional models, the diversity of body shapes and backgrounds in the dataset is limited. The main challenges of this dataset are the following: (1) we have limited amount of articles that have fit or shape label, (2) the label distribution is very imbalanced, with some of the fit and shape combinations having only a handful of samples, (3) information about intricate and elaborate physical fit attributes of garments is compressed in a few categorical labels, which leads to a highly complex task of annotation often resulting in noisy labels. We refer to our data variants in the following sections with the following naming convention;~\textbf{D} or~\textbf{J}, meaning Dresses or Jeans category followed by~\textbf{P} or~\textbf{M}, referring to packshot or model images and finally~\textbf{F} and/or~\textbf{S} to signal fit or shape labels. 

\subsection{Conditional Generation} \label{sec:conditional_generation} 

\begin{figure}[t!]
\centering
  \includegraphics[width=0.9\linewidth]{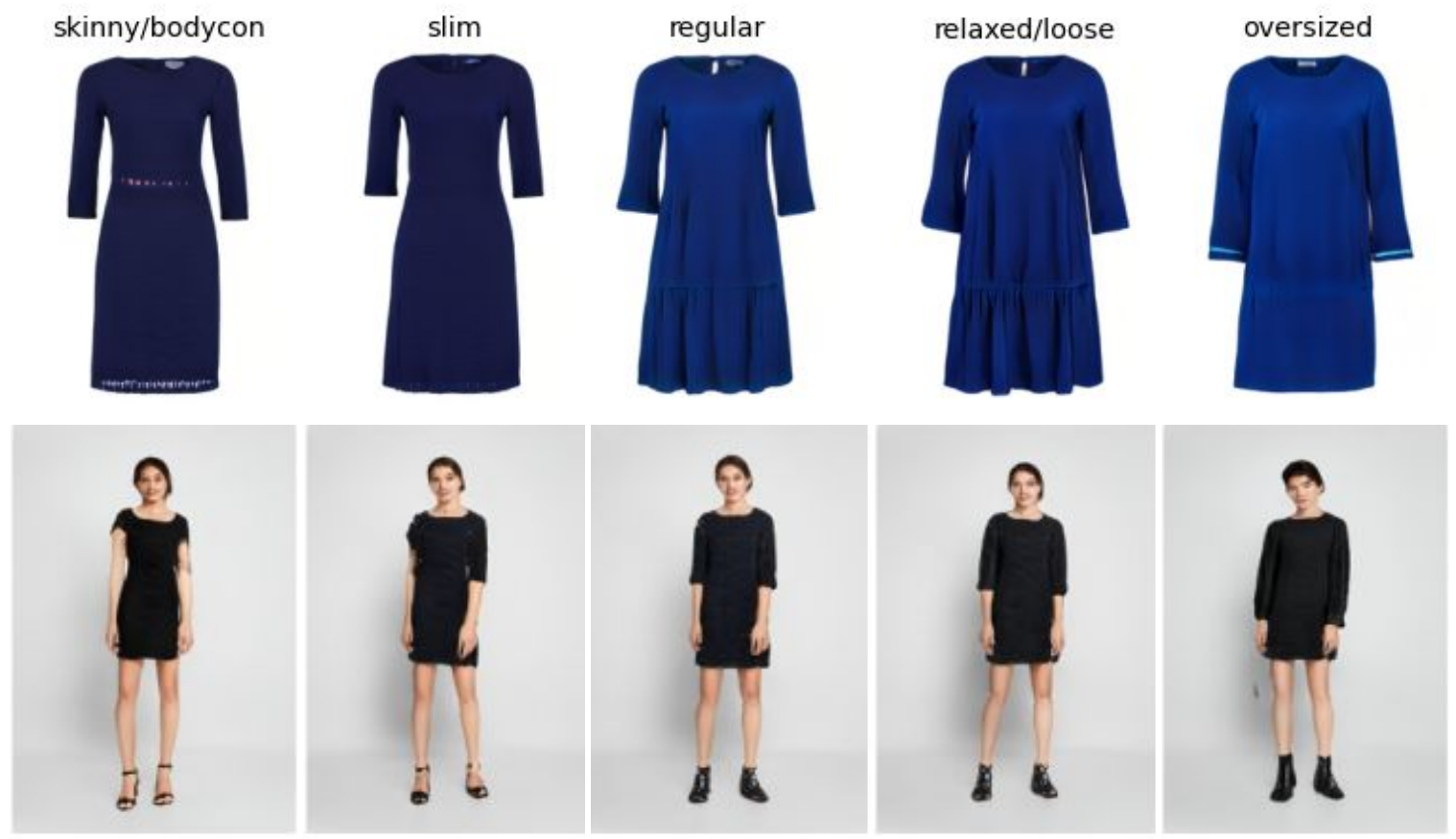}
  \caption{\textbf{Conditional fit generation}. A random input vector is conditioned on different fit levels. The generated images show underlying entanglements in the data, such as oversized dresses being lighter color or styled with boots.}
  \vspace{-.2cm}
\label{fig:conditional_dress_fit}
\end{figure}

We train a supervised FitGAN described in~\blackautoref{sec:method} separately on packshot and model images of dresses and jeans. We evaluate whether the output of the GAN is photo-realistic as well as whether the fit and shape attributes are synthesized realistically. To evaluate the network's outputs, we calculate standard GAN metrics, which can be found in~\blackautoref{tab:gan_metrics}. These metrics mostly focus on the fidelity and variety of the outputs, i.e. how photo-realistic and diverse the images are. Additionally, we inspect the produced output visually to asses the quality of the generated fit and shape aspects of the garments, as in~\blackautoref{fig:conditional_dress_fit}.

In most of the inspected network's outputs we notice that the network learns correlations present in the training data distribution. For example, as seen in~\blackautoref{fig:conditional_dress_fit}, as the fit of a dress is conditioned to be more loose, the color lightens. We can also see that oversized dress is styled with boots, whereas tighter fits are styled with pumps. As such implicit entanglements are present in the data distribution, we conclude that purely conditioning the network on fit and shape label is not enough to gain control over these elements. 

\begin{figure}[t!]
\centering
  \includegraphics[width=0.95\linewidth]{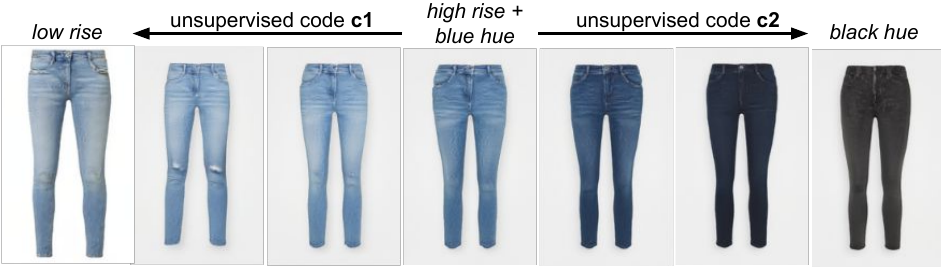}
 
  \caption{\textbf{Semi-supervised shape generation}. For a generated sample of jeans (middle) conditioned on body-hugging shape, we vary the unsupervised conditioning codes, which learn to represent the rise height (left) and color of the jeans (right).}
   \vspace{-0.5cm}
\label{fig:semi-supervised-jeans}
\end{figure}

We further experiment with the semi-supervised conditioning style described in~\autoref{sec:method} to gain more control over the implicit entanglements in the dataset by adding two continuous unsupervised labels to the conditioning. As shown in  \autoref{fig:semi-supervised-jeans}, a pair of jeans conditioned in a supervised manner to be of body-hugging shape, can be further modified in its rise height and color hue by varying the continuous codes, while the underlying body-hugging shape is persistent. We also note that for semi-supervised experiments we extended the generator mapping depth to 8 layers (instead of 2), which seems to have a positive impact on the FID score and Precision.

\begin{table}[b!]
\caption{\label{tab:gan_metrics}\textbf{Metric evaluation.} We evaluate our results with the standard GAN metrics: Inception Score (IS)~\cite{salimans2016improved}, Frechét Inception Distance (FID)~\cite{heusel2017gans} and Precision and Recall~\cite{kynkaanniemi2019improved}. Networks trained on images of dresses tend to perform better, which might be due to larger amount of available training data. Initial S- stands for supervised model, SS- indicates semi-supervised models.}
\vspace{-0.3cm}
\begin{center}
\begin{tabular}{l|rrrr}
Variant &   IS $\uparrow$ &   FID $\downarrow$ &    Precision  $\uparrow$ &   Recall $\uparrow$ \\
\hline
S-D-P-F &       
    $3.250$ &         
    $7.550$ &             
    $0.761$ &                 
    $0.177$  \\
S-D-M-F &       
    $2.403$ &         
    $12.285$ &             
    $0.582$ &                 
    $0.140$  \\
S-J-P-S &       
    $1.658$ &         
    $20.675$ &            
    $0.513$ &                 
    $0.003$  \\
SS-J-P-S &       
    $1.422$ &         
    $\textbf{4.716}$ &            
    $\textbf{0.776}$ &                 
    $0.126$  \\
S-J-M-S &       
    $2.380$ &         
    $21.728$ &            
    $0.606$ &                 
    $\textbf{0.184}$  \\
S-D-P-F+S & 
    $\textbf{3.651}$ &
    $9.733$ & 
    $0.698$ &
    $0.171$ \\
\hline

\end{tabular}
\end{center}
\vspace{-0.4cm}
\end{table}

\subsection{Latent Space Analysis} \label{sec:latent_space}

\begin{figure}[b!]
\centering
  \includegraphics[width=1\linewidth]{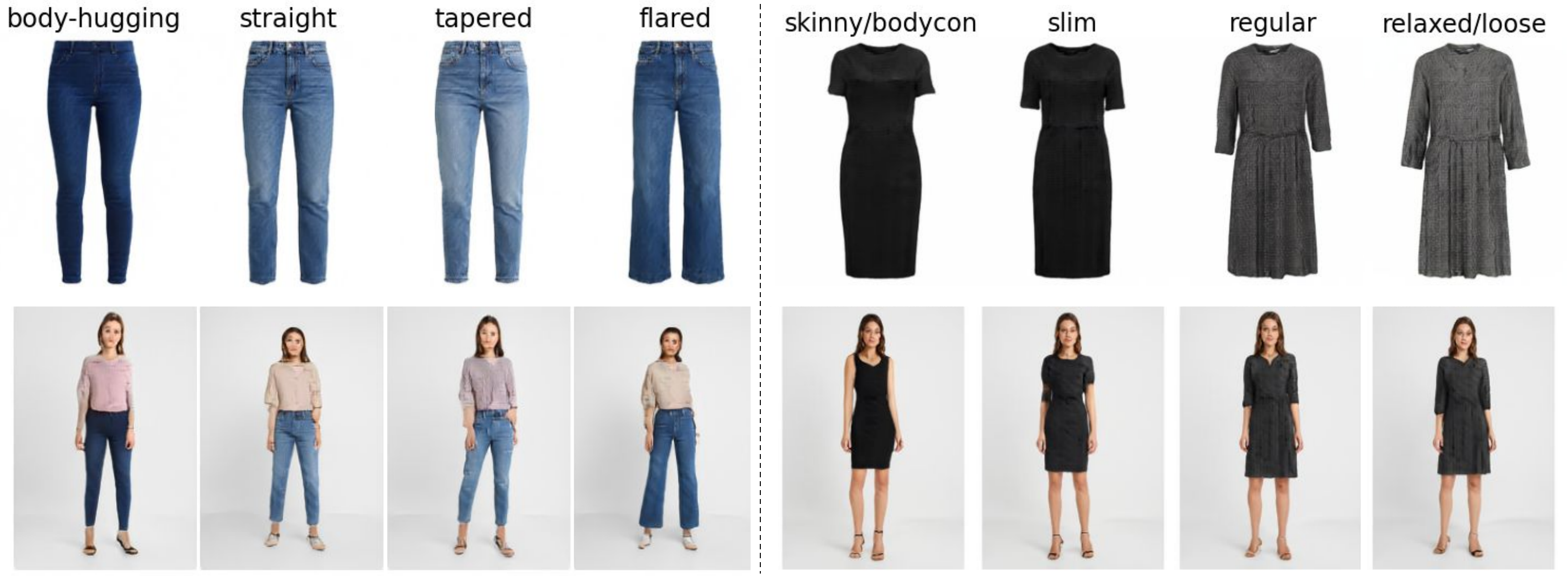}
   \caption{\textbf{Average class representations}. The images display what an average fit and shape looks like in the latent space and surface some entanglements in the dataset, e.g.: looser dress is on average lighter in color with longer sleeves.}
\label{fig:mean_class_representations}
\end{figure}

We study the learnt intermediate latent space \textbf{W} of the supervised FitGAN and how the fit information is encoded in it. More specifically, we investigate whether the fit and shape are encoded in a semantically meaningful way. We compute a class representation as the mean latent vector ${\mu_{w_c}}$ for a class $c$. We obtain this by averaging the latent codes $w$ of 10.000 random input vectors conditioned on the class $c$. ~\blackautoref{fig:mean_class_representations} shows the generated example from mean class representations. We can see that the network correctly encodes fit and shape and also what an average garment with such fit or shape looks like -- for example, looser garments tend to have longer sleeve length and lighter color, and body-hugging jeans tend to be darker.
One difference between the fit and shape labels is that fit can be ordered into fit levels, where as there is no natural ordinality for shape. In order to validate whether the network's latent space is able to encode this ordinality, we compute the euclidean distances between the mean latent vector representations ${\mu_{w_c}}$. We observe, that Bodycon fit has a distance of 2.7 to Skinny fit, 3.9 to Regular fit, 4.4 to Loose fit and 5.3 to Oversized fit. This signifies that the implicit order of the fit is represented in the latent space, with an average bodycon fit closer to skinny fit than to regular and so on. This behaviour is consistent across all fit classes. For full results please refer to Supplementary Materials. 
We note here that these results are produced with a network which was trained to be conditioned solely on the fit attribute in one-hot-encoded form. Therefore we argue that the ordinality is not explicitly given, but rather implicitly understood by the model. On the other hand, in a network that was conditioned only on one-hot-encoded shape attribute, each class representation ${\mu_{w_c}}$ is encoded at approximately the same distance from all others. \\

\begin{figure}[b!]
     \centering
     \begin{subfigure}[t]{0.49\textwidth}
         \centering
         \includegraphics[width=\textwidth]{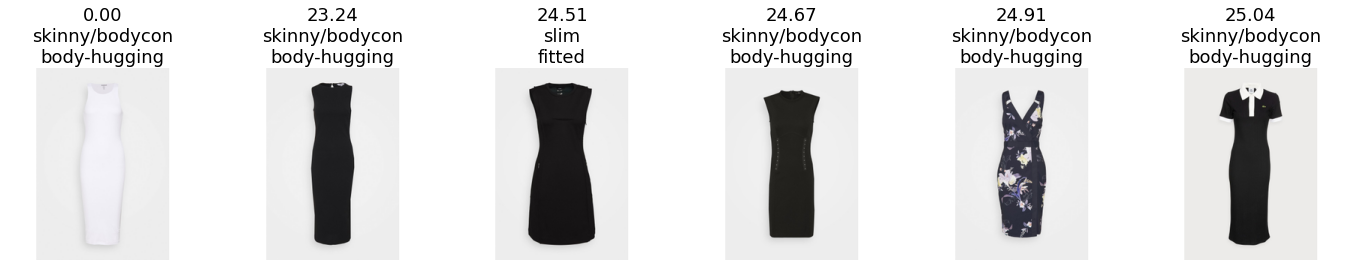}
     \end{subfigure}
     \vfill
          \begin{subfigure}[t]{0.49\textwidth}
         \centering
         \includegraphics[width=\textwidth]{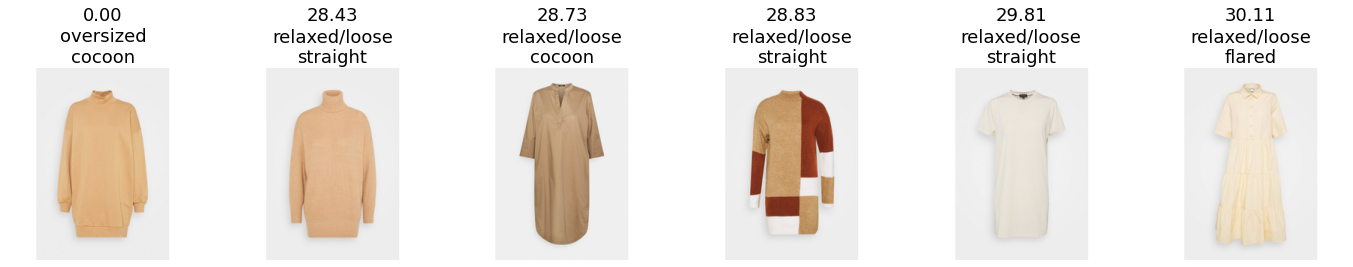}
     \end{subfigure}
    
    \caption{\textbf{Finding similar articles.} For a query image (most left) we find articles that are closest in the latent space using Euclidean distance. The distance, article's fit and shape label are displayed above each image. Retrieved articles have similar fit and shape as well as other visual characteristics.}
    \label{fig:similar_articles}
\end{figure}

We further explore the use of the latent space to find articles in similar fits and shapes. As shown previously, the latent space is able to semantically organize fit of the garments. Therefore our assumption is that articles with similar fits and shapes are located close to each other in the latent space. \blackautoref{fig:similar_articles} shows an example of fit and shape aware similarity search, where based on a query article, the articles with the smallest euclidean distance are similar in both fit and shape, as well as other visual characteristics. We note here that similar images are selected from a limited amount of test articles that have been projected to the latent space; the potential application could find even more similar images from a larger catalogue of articles. For more experimentation with similarity search, please refer to the Supplementary Materials.

\subsection{Real Image Projection}

Apart from generating novel content, we are also interested in projecting existing garments into the latent space of the GAN and ultimately, controlling their appearance by the conditioning. The projection of real images into the latent space is an active and challenging area of research, where the state of the art methods fail to reconstruct the real garment details\cite{lewis2021tryongan}. We use the most common technique, as implemented by~\cite{karras2020training}, that finds the optimal latent code $w$ which produces an image closest to the real image, minimizing the perceptual distance.

Once we find the optimal latent code $w$ for a real image, we transform it to a different fit and shape by using a style mixing approach introduced in~\cite{karras2019style}. We use this approach, as in the current architecture it is not possible to directly condition the latent code $w$ with a class label $c$, as the conditioning of the generated image happens in the input latent space~$Z$. Therefore, we use the mean class representation vectors introduced in~\blackautoref{sec:latent_space} and mix the projected style code $w$ of the image with the mean style code of the target class at hand-picked layers of the generator.

\begin{figure}[t!]
     \centering
     \begin{subfigure}{0.49\textwidth}
         \centering
         \includegraphics[width=\linewidth]{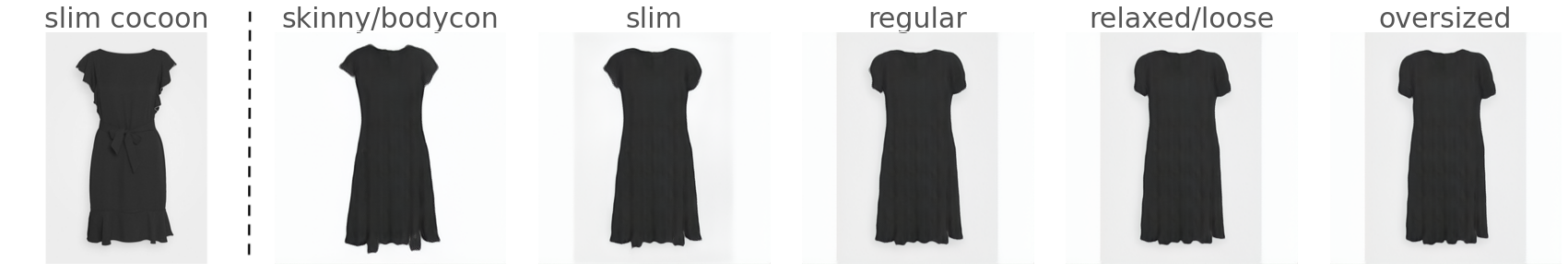}
     \end{subfigure}
     \vfill
     \begin{subfigure}{0.49\textwidth}
         \centering
         \includegraphics[width=\linewidth]{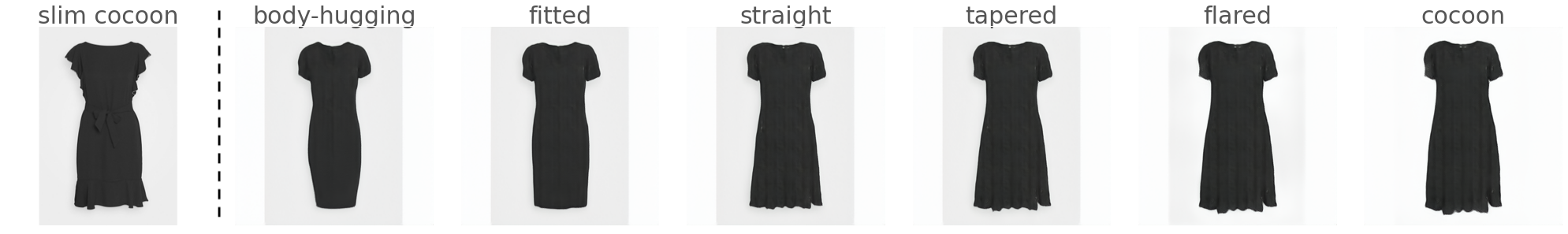}
     \end{subfigure}
     
      \caption{\textbf{Real image projection}. A real image (most left) is projected into the latent space of the GAN and reconstructed by the generator while conditioned on various fits (top) and shapes (bottom). Some details such as sleeve structures and belt are lost during reconstruction.}
      \vspace{-0.4cm}
\label{fig:real_image_projection}
\end{figure}

As we can see in~\blackautoref{fig:real_image_projection}, the projected version of the article does change its appearance by applying the style of other fit and shape classes, with fit having a more visible impact on the image than shape. However, the image that is projected to the latent space $W$ and subsequently reconstructed by the generator back to the image space loses some of its details. 
This suggests that the latent space lacks the ability to encode such details and the generator collapses to a more common node with simple sleeves. For more details on latent space disentanglement, using pre-trained article representations, please see Supplementary Materials.

\subsection{Evaluation}

\subsubsection{Domain Expert Annotation}
Although state-of-the-art metrics to evaluate GANs can provide guidance on how realistic the output is, the main goal of FitGAN is to also realistically visualize fit and shape of the garment. We have used an expert in the domain of article fit to annotate real and generated images of dresses with the appropriate fit and shape attribute. We consider the domain expert's annotations ground truth and compare the accuracy against target generator condition to evaluate the truthful visualization of fit and shape. As shown in \blackautoref{tab:evaluation_metrics}, we found that our expert agreed with the training annotations on 73\% of real images. As the original training dataset is annotated by trained semi-professionals, we expect this slight deviation from the expert's opinion. We compare it with the accuracy between expert's opinion and the target condition label of generated images, which is 47\% and 56\% for fit and shape attributes of the generated garments respectively. Given that a random generator would be able to generate the right fit and shape 20\% and 16.6\% of times respectively, we find this result to show that our FitGAN is able to produce realistic fit and shape of garments. Please refer to Supplementary Materials for detailed method of annotation.

To validate our assumption on the value of introducing explicit fit and shape control mechanism in the generator process, we also compare our work to the Outfit Renderer~\cite{yildirim2019generating}, which generates a human model in a given dress, taking as input the dress packshot image and a target pose. This method does not consider any explicit fit and shape information about the garment and we hypothesize that although the garments might appear visually realistic, the actual fit and shape of the garment might not be preserved on the generated human model. We find that our domain expert agrees with the original article fit annotation around 41\% of times, and 40\% of times with the shape annotation when seeing the generated outfit image. We ask our domain expert to also rate how realistic an image is, details of which can be found in the Supplementary Materials.

\subsubsection{Automated Annotation}
As manual data annotation is a time-consuming and laborious process, we use our auxiliary supervised classification network introduced in ~\blackautoref{sec:evaluation-model} to classify fit and shape attributes and evaluate the results on a larger scale.
We follow a similar strategy as for the domain expert annotation with larger dataset. We find the classifier reaches similar accuracy to the performance of our domain expert annotator on real images. On generated images, comparing the classifier predictions with the target fit and shape condition, the classifier reaches an accuracy of 54\% and 62\% on fit and shape respectively. Model trained on human images is less confident on both real images and fake images, as shown in~\blackautoref{tab:evaluation_metrics}.

\begin{table}[t!]
\caption{\label{tab:evaluation_metrics}\textbf{Accuracy of domain expert and automated classifier annotations on Women's Dresses.} We compare a FitGAN variant with Outfit Renderer~\cite{yildirim2019generating} based on the accuracy of the domain expert annotations. Using annotations from supervised classifier, we evaluate two FitGAN variants, where target generator condition is considered ground truth for fake images.}
\vspace{-0.3cm}
\centering
\begin{tabular}{l|l|c|c|c|c}
\multicolumn{2}{c|}{} & \multicolumn{2}{c|}{Fake Images} & \multicolumn{2}{c}{Real Images}\\
\hline
Annotator & Model & Fit & Shape & Fit & Shape\\
\hline
Expert & D-P-FS & 0.474 & 0.564 & 0.730 & 0.730 \\
Expert & Outfit & 0.412 & 0.404 & 0.300 & 0.400 \\ 
\hline
Classifier & D-P-FS & 0.538 & 0.618 & 0.636 & 0.707 \\ 
Classifier & D-M-FS & 0.489 & 0.524 & 0.604 & 0.670  \\ 
\end{tabular}
\vspace{-0.3cm}
\end{table}

%% file: sections/05_conclusion.tex
We introduced a generative framework FitGAN and demonstrated  through extensive experimentations with real-world data at scale, how such a flexible generative approach can be leveraged to tackle the challenging problem of size and fit in fashion. We demonstrated that FitGAN is capable of explicitly learning internal representations of article specific fit and shape properties built on visual clues from 2D images. We evaluated FitGAN in different scenarios against highly trained fashion expert annotations, pool of regularly trained semi-expert human annotations, and supervised machine classification approaches. We showed solid results on condition the output of a generative network on a limited dataset of categorical fit and shape labels, and explored semi-supervised conditionings at scale, for achieving valuable realistic generation and transformation of an article’s fit and shape in the 2D image space. In order to tackle some of the limitations of the current method, future work will focus on experimenting with different generative models and exploring further semi and self-supervised approaches. This work enables promising future work on improving the latent space to best encode intricate details of the garments, for complex patterns to achieve realistic projection of real images, as well as achieving better disentanglement of other article attributes such as color and pattern, and ensuring that a. the article's identity is kept throughout conditioning on different fits and shapes, and b. the implicit impact of those attributes are incorporated in the perceived fit and shape.